\documentclass[conference]{IEEEtran}
\IEEEoverridecommandlockouts
\usepackage{cite}
\usepackage{amsmath,amssymb,amsfonts}
\usepackage{algorithmic}
\usepackage{graphicx}
\usepackage{textcomp}
\usepackage{adjustbox}
\usepackage{subcaption}
\usepackage{multirow}

\usepackage{array} 
\usepackage{caption} 

\usepackage{xcolor}
\usepackage[font=footnotesize,belowskip=-2pt]{caption}
\usepackage[normalem]{ulem}
\useunder{\uline}{\ul}{}
\def\BibTeX{{\rm B\kern-.05em{\sc i\kern-.025em b}\kern-.08em
    T\kern-.1667em\lower.7ex\hbox{E}\kern-.125emX}}
\usepackage[letterpaper, top=0.78in, left=0.68in, right=0.68in, bottom=1.1in]{geometry}

\begin{document}

\title{LLMcap: Large Language Model for Unsupervised PCAP Failure Detection}

\author{\IEEEauthorblockN{Łukasz Tulczyjew}
\IEEEauthorblockA{
\textit{B-Yond}\\
Gliwice, Poland \\
lukasz.tulczyjew@b-yond.com}
\and
\IEEEauthorblockN{Kinan Jarrah}
\IEEEauthorblockA{
\textit{B-Yond}\\
Belgrade, Serbia \\
kinan.jarrah@b-yond.com}
\and
\IEEEauthorblockN{Charles Abondo}
\IEEEauthorblockA{
\textit{B-Yond}\\
Montreal, Canada \\
charles.abondo@b-yond.com}
\and
\IEEEauthorblockN{Dina Bennett}
\IEEEauthorblockA{
\textit{B-Yond}\\
Montreal, Canada \\
dina.bennett@b-yond.com}
\and
\IEEEauthorblockN{Nathanael Weill}
\IEEEauthorblockA{
\textit{B-Yond}\\
Montreal, Canada \\
nathanael.weill@b-yond.com}
}

\maketitle

\begin{abstract}
The integration of advanced technologies into telecommunication networks complicates troubleshooting, posing challenges for manual error identification in Packet Capture (PCAP) data. This manual approach, requiring substantial resources, becomes impractical at larger scales. Machine learning (ML) methods offer alternatives, but the scarcity of labeled data limits accuracy. In this study, we propose a self-supervised, large language model-based (LLMcap) method for PCAP failure detection. LLMcap, leveraging language-learning abilities, employs masked language modeling to learn grammar, context, and structure. Tested rigorously on various PCAPs, it demonstrates high accuracy despite the absence of labeled data during training, presenting a promising solution for efficient network analysis.
\end{abstract}

\begin{IEEEkeywords}
Network troubleshooting, Packet Capture Analysis, Self-Supervised Learning, Large Language Model, Network Quality of Service, Network Performance 
\end{IEEEkeywords}

\section{Introduction}\label{section_intro}
In the dynamic landscape of today's telecommunications industry, the proliferation of sophisticated technologies and the intricate web of network components have introduced unprecedented challenges to the task of network troubleshooting. The sheer volume of data and messages exchanged between these components, coupled with the need for swift and accurate fault detection, demands innovative solutions that transcend the limitations of traditional methodologies. 

LLMcap is a self-supervised large language model (LLM)-based method designed to revolutionize fault detection in telecommunication networks. In an era where Mean Time to Detect, Mean Time to Repair, customer experience, and the detection of hidden fault types are paramount, LLMcap is an adaptable solution, providing a change in basic assumptions in how we approach network fault detection.    

LLMcap examines the extensive data and messages exchanged between network components and identifies abnormalities and failures. It achieves this without requiring extensive domain knowledge or a deep understanding of diverse protocols. Its implementation promises to significantly improve both the Mean Time to Detect and Mean Time to Repair for telecom operators, ultimately enhancing the overall customer experience. Furthermore, its capability to detect hidden failure types and measure abnormalities in message sequences and timing adds crucial dimensions to ensuring the reliability and efficiency of the telecommunication network. 

This paper introduces an innovative LLM-based system designed for the efficient detection of failed PCAP files. The key contributions of our research include:

\begin{itemize}
    \itemsep0em
    \item Utilization of state-of-the-art LLMs to acquire a deep understanding of the representation and grammar of successful PCAP files. 
    \item Eliminating the need for labeled data by using self-supervised LLMcap, making the solution adaptable to various use cases. 
    \item High accuracy in failure detection within PCAP data within our experimental results. 
    \item Accurate localization of failures in PCAP providing valuable insights for analysis. 
    \item Adaptive and flexible methodology that allows each building block to be replaced with an equivalent approach, enhancing versatility and customization. 
    \item A hypothesis that aggressive reduction of PCAP data into a dictionary format, with specific fields/entities, prior to training, does not result in a statistically significant loss in the quality of failure detection. 
    \item A new PCAP representation in the form of a key-value dictionary to train the LLM to reduce the problem's dimensionality. This representation contributes to the overall efficiency of our approach. 
\end{itemize}

\section{Related Work}\label{section_related_work}

Packet capture (PCAP) data plays a crucial role in network troubleshooting and analysis. Prior work in the field has explored various strategies for detecting failures within PCAP-captured network traffic. Studies have focused on identifying abnormal traffic patterns, indicative of network issues or security threats. Traditional approaches have been heavily reliant on manual inspection and rule-based systems, which are time-consuming and may not scale with the increasing volume of network traffic \cite{b1}. Researchers have explored automated methods, including statistical analysis and simple heuristics, to improve the efficiency of failure detection in PCAP files \cite{b2}. 

The application of machine learning (ML) techniques to PCAP data is a relatively new field that has gained traction due to its potential to automate and enhance the accuracy of network analysis. Several studies have employed supervised learning methods, utilizing labeled datasets to train models for classifying network traffic and identifying malware traffic \cite{b3}. However, the challenge of obtaining a sufficiently large and diverse labeled dataset has been a significant barrier, often limiting the effectiveness of these ML models \cite{b4}. Unsupervised learning techniques, such as clustering and outlier detection, have been investigated as alternatives to address the scarcity of labeled data \cite{b5}. The main drawback of machine learning approaches is that they rely on handcrafted features and labeled data.  

Large language models (LLMs) have recently emerged as powerful tools for anomaly detection across various domains. Their capacity to understand context and infer patterns within data makes them well-suited for identifying irregularities \cite{b6}, \cite{b7}. In the realm of network security, LLMs have been leveraged to parse and analyze network logs and textual data, demonstrating an ability to detect sophisticated threats that conventional methods might miss. The concept of self-supervision in LLMs, where models learn to identify anomalies by inferring the underlying data structure, has shown promise in enhancing detection capabilities without the need for labeled examples \cite{b8}. 

In conclusion, while traditional and ML-based methods for PCAP failure detection have laid the groundwork, the integration of LLMs in this domain, as proposed by our method LLMcap, has the potential to redefine the benchmarks of accuracy and efficiency in the field. 

\section{Problem Formulation}\label{section_problem_formulation}

\begin{figure}
\centerline{\includegraphics[width=\columnwidth]{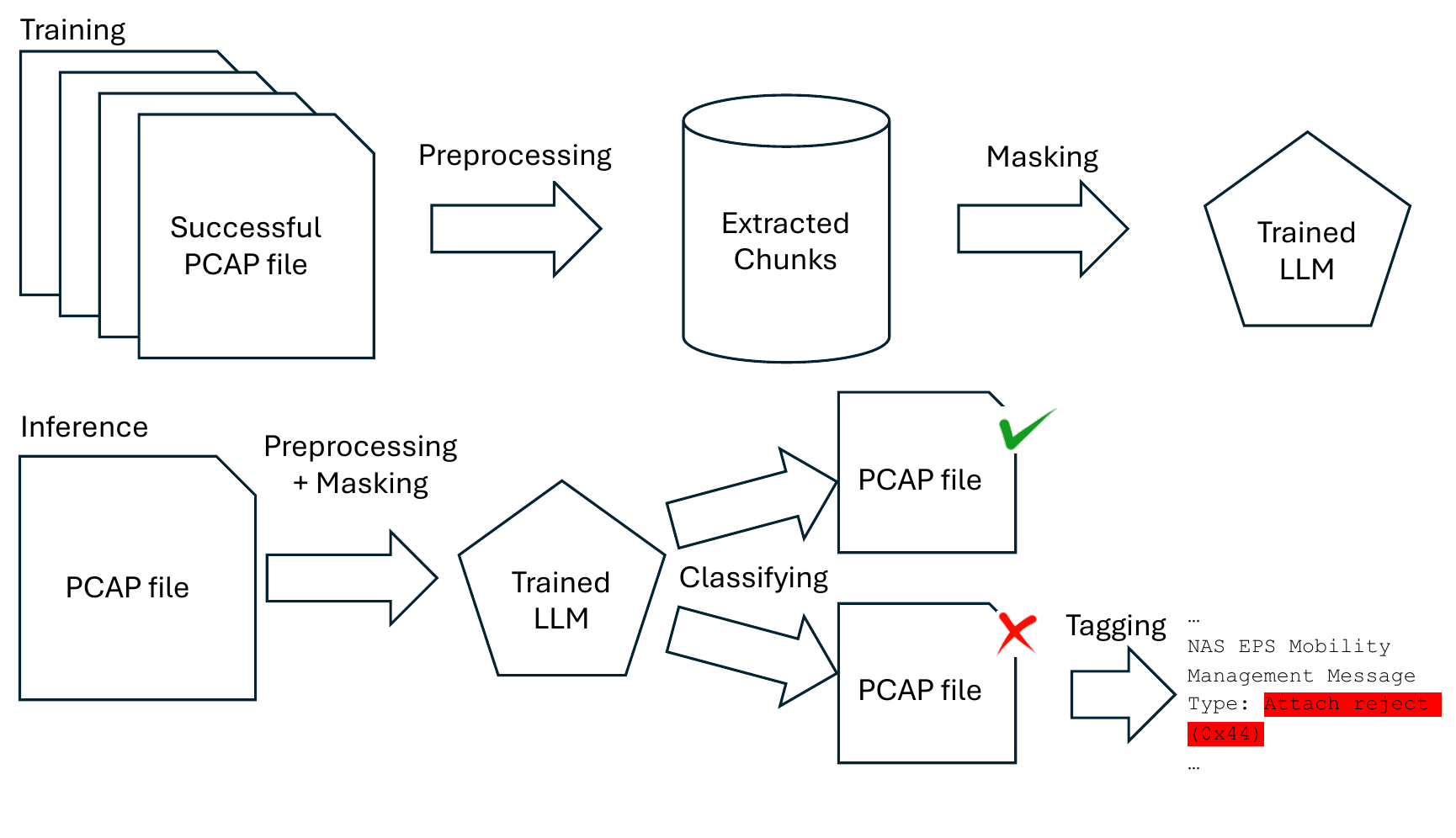}}
\caption{The overall strategy for training and inference using LLMcap.}
\label{fig1}
\end{figure}

In this study, we address the challenge of PCAP failure detection by employing the Mask Language Model (MLM) strategy. Our definition of failure is a non-successful call. We make a clear distinction between the broader concept of error which can be any form of feedback from the network elements, and failure, where the intended initiated call didn’t succeed. We hypothesize that an MLM approach is well-suited to identify the occurrence of failures in each PCAP file. In the latter, the model can effectively pinpoint the starting point for the operator’s investigation by using the reconstruction error (tagging), as illustrated in Figure \ref{fig1}. 


In the context of PCAP data, a single instance can be viewed as a sequence of consecutive character strings, (called tokens). The MLM approach utilizes unannotated text and randomly replaces a portion of words with a [MASK] token \cite{b8}, \cite{b9}. The training objective for the model is to accurately reconstruct the intentionally hidden words based on the context and word semantics in the input sequence (Figure \ref{fig1}, training). The training phase comprises two steps. The first step is to preprocess the PCAP file to obtain a set of chunks, and the second step is to randomly mask tokens for the model to predict. Once the model is trained, the inference procedure follows the same path encompassing the preprocessing and masking steps (Figure \ref{fig1}, inference). 

Given a set of $N$ tokens in an input sequence $T={t_i}^{N}_{i=1}$, a random subset of $M$ tokens undergoes the masking procedure.

For each reconstructed word, the Negative Log-Likelihood (NLL, Equation \ref{eq1}) loss is computed:
\begin{equation}
    NLL=-\frac{1}{M}\sum^{M}_{i=1}\sum^{C}_{c=1}y^{c}_{i}log(a^{c}_{i}),
\label{eq1}
\end{equation}

where, $y^{c}_{i}$ is the one-hot encoded label, which can be either equal to 0 or 1, for the correct token. Furthermore, $a^{c}_{i}$ denotes the predicted probability for the corresponding true token when $y^{c}_{i}$ is set to 1.

It is worth noting that $a_i$ vector undergoes the Softmax activation function before, thus its sum corresponds to the value of 1, where each entry represents the probability of selecting the 
$c$-th token.

We define the Softmax activation function as: 
\begin{equation}
    \sigma=\frac{e^{x_i}}{\sum^{C}_{c=1}e^{x_c}},
\label{eq2}
\end{equation}
where the $\sigma$ symbol defines the Softmax nonlinearity function.

The formula in Equation \ref{eq1} denotes the cost function, i.e., loss, which is incorporated in the training phase of the LLM. 

\section{Proposed System/Model}\label{section_proposed_model}

This segment outlines the proposed approach for detecting failures in PCAPs. The process is broken down into distinct parts. Data collection is covered in Subsection \ref{sub_4_1}. Subsection \ref{sub_4_2} focuses on the PCAP parser and preprocessing steps. Subsection \ref{sub_4_3} delves into the training strategy for LLM and MLM. The concluding Subsection \ref{sub_4_4} introduces the Failure Detection algorithm (FDA). 

\subsection{Data Collection}\label{sub_4_1}

This study focused on Packet Capture (PCAP) files for their adherence to standardized formats, capturing detailed message information consistently across network components to provide a holistic view. Our dataset comprises 1.263 gigabytes of PCAP files, spread across 3,318 individual captures, highlighting diverse scenarios within Voice over LTE (VoLTE), Voice over New Radio (VoNR), and Connectivity services such as Attach, Registration, and PDU establishments. Our dataset features 1,675 PCAP files for VoLTE scenarios, 1,246 files for Connectivity scenarios, and 397 files for VoNR scenarios. These captures were obtained from end-to-end (e2e) network interfaces and cover a comprehensive representation of protocols including S1AP, NGAP, SIP, GTPV2, DIAMETER, HTTP2, and PFCP.

These PCAP files were collected from a laboratory network designed to mimic real-world network conditions, offering a broad spectrum of call flow scenarios such as VoLTE-to-VoLTE calls, VoLTE-to-fixed network calls, VoNR-to-VoNR calls, and complex interactions within 4G and 5G Non-Standalone (NSA) combo networks. This diversity ensures our dataset's applicability and robustness across different network environments, with scenarios covering end-to-end calls, registration processes, session establishments, context creations, and charging flows.

Our dataset is categorized into successful and failure groups, focusing on detailed failure types, including explicit error failures (error codes and timeouts) and implicit error failures. Notably, the data exhibits varied failure rates, with a 22\% failure rate for VoLTE data, an 83\% failure rate for VoNR data, and a 21\% failure rate for Connectivity flows. This categorization distinguishes successful scenarios, representing calls without failed flows, from those containing failed end-to-end calls. A PCAP is deemed to contain failure if at least one call flow within it is unsuccessful.

To ensure the integrity and reliability of our training data, Subject Matter Experts (SMEs) curated the dataset and annotated the dataset using our proprietary software AGILITY \cite{b10}. This crucial step guarantees high-quality results, providing the ground truth for our models. Including only successful samples, i.e., PCAPs without failures or errors, in our dataset as shown in Table \ref{tab1}, we aim to efficiently train our model with a rich and diverse dataset, highlighting our approach's innovation and its potential for broad application in real-world telecommunication networks.

\begin{table}[]
\centering
\begin{tabular}{|c|c|c|}
\hline
Service Name           & Successful PCAPs & Failed PCAPs \\ \hline
VoLTE                  & 1311                       & 364                    \\
VoNR                   & 66                         & 331                    \\
4G-5G NSA Connectivity & 982                        & 264                    \\ \hline
\end{tabular}
\caption{Distribution of the dataset across different services.}
\label{tab1}
\end{table}

\subsection{PCAP Parser and Preprocessor}\label{sub_4_2}

In this section, we outline the four steps used to process PCAP files for MLM training and inference. The four steps: Parsing, Data Sanitization, Chunking, and Masking, are described in greater detail below. Figure \ref{fig3} provides an example of input and output resulting from this process. 

\begin{figure}[h]
    \centering
    \begin{adjustbox}{width=\columnwidth}  
        \texttt{\begin{tabular}{|l|l|}
            \hline
            S1 Application Protocol & S1 [MASK] Protocol\\ 
            S1AP-PDU: initiatingMessage (0) & S1AP-PDU: [MASK] (0) \\
            initiatingMessage & [MASK] \\
            procedureCode: id-initialUEMessage (12) & procedureCode: id-[MASK] (12) \\ 
            criticality: ignore (1) & criticality: [MASK] (1) \\ 
            value & [MASK] \\ 
            InitialUEMessage & InitialUEMessage \\
            protocolIEs: 6 items & [MASK]: 6 items \\ 
            Item 0: id-eNB-UE-S1AP-ID & [MASK] 0: [MASK] \\ 
            ProtocolIE-Field & ProtocolIE-Field \\ 
            id: id-eNB-UE-S1AP-ID (8) & id: id-eNB-UE-S1AP-ID (8) \\
            criticality: reject (0) & criticality: [MASK] (0) \\
            value & value \\
            ENB-UE-S1AP-ID: 3039369 & ENB-UE-S1AP-ID: [REDACTED] \\ \hline
        \end{tabular}}
    \end{adjustbox}
    \caption{Example of formatted original and masked input to MLM training of LLM model.}
    \label{fig3}
\end{figure}

\subsubsection{Parsing}\label{sub_sub_4_2_1}

Our method explored two distinct approaches for extracting the PCAP representation. The first involved extracting all available data as text from the PCAP file, offering a comprehensive view of the sample but increasing the computational burden during training. This enables treating a set of PCAPs as an extensive corpus, adopted during MLM training.

Conversely, the second method extracted a set of packet-separated fields from the PCAP as a dictionary. This facilitates substantial data reduction by incorporating only selected fields based on a priori or SME knowledge. 

The extraction was performed using \emph{tshark v4.0.7} \cite{b11}, employing the “\emph{text}” flag for text extraction and the “\emph{pdml}” flag for PDML extraction. Protocol filters (S1AP, NGAP, SIP, GTPV2, DIAMETER, HTTP2, PFCP, LCSAP, ULP, DNS, ICMP, TCP, RTP, F1AP, UDP, and HTTP) are applied. 

In the PDML scenario, before obtaining dictionary representation, we converted the PCAP into a tabular format. This conversion allows structured data operations, indexing, and sorting, enhances readability and provides flexibility in the output. 

\subsubsection{Data Sanitization}\label{sub_sub_4_2_2}

PCAP files inherently display variability, posing challenges for accurate prediction due to differing data elements. Additionally, user-related information in PCAP files, such as IP and MAC addresses, is sensitive, necessitating precautions to prevent the model from learning and potentially exposing this data. To address these concerns, we systematically removed user-specific details, by employing regular expressions. Sensitive information, like IP and MAC addresses, was replaced with generic tokens, the [REDACTED] symbol---to maintain the integrity of the data while safeguarding user privacy within PCAP files. 

\subsubsection{Chunking}\label{sub_sub_4_2_3}

At this stage, the obtained file is very variable in size. In this study we standardized the context length to a relatively small size (set to 64), to expedite training and streamline model iteration. This was achieved by tokenizing the text and dividing it into equal-sized chunks with the designed number of tokens. For the last chunk, we applied a padding operation, to preserve the constant dimensionality to ensure a standardized context length.  

\subsubsection{Masking}\label{sub_sub_4_2_4}
To train the LLM, we employed a masking approach wherein a set of tokens is randomly masked. We randomly selected 20\% of tokens and replaced them with [MASK] symbols. The dataset was structured to include chunks with both masked tokens and non-masked tokens. 

\subsection{Large Language Model Training}\label{sub_4_3}

In this section, we provide details of our large language model (LLM) training process, highlighting our choice of the Transformer-based DistilBERT model for masked language modeling \cite{b12}. Our approach follows the MLM technical solution defined in \cite{b8}, with two modifications.  Firstly, we only use the [MASK] token word for replacement in contrast to additional random word replacement in \cite{b8}), and secondly, we set the masking probability to 20\%. We hypothesize that the 5\% increase from \cite{b8}, improves the LLM generalization capabilities, enhances feature extraction by emphasizing more salient keywords within the highly specialized PCAP representation, and improves robustness of the model during the inference. Furthermore, we omitted the next sentence prediction method defined in \cite{b8}, reserving it for future exploration, and focus solely on MLM in this study. For clarity, we include here relevant implementation details. 

\begin{table*}[]
\centering
\begin{tabular}{|>{\centering\arraybackslash}m{2.2cm}|>{\centering\arraybackslash}m{2.4cm}|>{\centering\arraybackslash}m{2.8cm}|>{\centering\arraybackslash}m{2cm}|>{\centering\arraybackslash}m{2cm}|>{\centering\arraybackslash}m{2.2cm}|}
\hline
\textbf{Model type} & \textbf{Encoder-Only (e.g., DistilBERT\cite{b12})} & \textbf{Encoder-Decoder (e.g., Transformer\cite{b14})} & \textbf{Decoder-Only (e.g., GPT\cite{b15})} & \textbf{Supervised ML} & \textbf{Unsupervised ML} \\ \hline
\textbf{Training approach} & Masking & Sequence-to-sequence & Next token prediction & Label-based & Pattern discovery \\ \hline
\textbf{Size and Compute} & Smaller, less compute & Larger, more compute & Large, resource-intensive & Varies & Generally less \\ \hline
\textbf{Data Requirement} & Efficient with unlabelled & Handles labeled and unlabelled, prefers diverse & Extensive datasets & Large, well-labeled & Unlabeled data \\ \hline
\textbf{Continuous Learning} & Possible with updates & Flexible, resource-dependent & Significant retraining & Limited by dataset & Adaptable without retraining \\ \hline
\textbf{Specific Characteristic} & Context understanding & Translation tasks  & Text generation & \multicolumn{2}{>{\centering\arraybackslash}m{4.4cm}|}{Data quality dependent, rely on data representation}  \\ \hline
\textbf{Data Encoding} & Minimal feature engineering & Sophisticated engineering needed & Minimal engineering & \multicolumn{2}{>{\centering\arraybackslash}m{4.4cm}|}{Initial feature engineering }  \\ \hline
\end{tabular}
\caption{Comparative Analysis of Machine Learning Model Characteristics}
\label{tableComp}
\end{table*}

\subsubsection{Model Selection: DistilBERT and Knowledge Distillation}\label{sub_sub_4_3_1}

To select the best architecture for the Failure Detection Algorithm, we thoroughly reviewed various LLM and Machine Learning approaches (see Table \ref{tableComp}). This table compares the characteristics of the original implementations for each class of Large Language Models (LLMs) and related machine learning approaches. Note that newer implementations may diverge from these characteristics, reflecting ongoing advancements and optimizations in the field. Based on our comparison, DistilBERT, with its encoder-only architecture, offers a balance of computational efficiency and high accuracy in contexts that require a deep understanding of data relationships, such as PCAP failure detection. Its efficiency in processing unlabelled data minimizes the need for extensive datasets and feature engineering, facilitating easier integration and scalability within network environments. Moreover, its potential for continuous learning with periodic updates enables adaptable anomaly detection in dynamic telecommunications networks. 

DistilBERT, a distilled version of the BERT model, employs knowledge distillation to reduce the model size while retaining 97\% of the language understanding capabilities of the original BERT model with a 40\% reduction in size. Additionally, DistilBERT operates at a 60\% faster speed, making it optimal for our training pipeline. 

In our DistilBERT architecture, the transformer encoder has 6 layers and 12 attention heads, and the intermediate layer size in the transformer encoder is set to 3072 units. For regularization and to prevent overfitting, the dropout rate in the fully connected layers was set to 0.1. The dimensionality of the encoder layers defaulted to 768. The maximum chunk size was set to 64. We applied the Gaussian Error Linear Unit (GELU) as the non-linearity. The vocabulary remained unchanged, with 30522 entries. 

\subsubsection{Loss Function and Training Configuration}\label{sub_sub_4_3_2}

To train the LLM, we employed Negative Log Likelihood (NLL) loss as the optimization criterion. The training process extends to a maximum of 200 epochs, ensuring the model converges to an optimal state. To prevent overfitting, we incorporated an early stopping mechanism by setting the patience hyperparameter to 12 epochs to halt training if there was no improvement in the cost function. 

\subsubsection{Input Token Masking and Shuffling}\label{sub_sub_4_3_3}

During training, we introduced a masking probability of 20\% to the input token sequence, enabling the model to learn contextual relationships by predicting masked tokens. To prevent the LLM from learning the sample order, we shuffled the dataset in each epoch, providing variability in the training data. 

\subsubsection{Experimental Setup}\label{sub_sub_4_3_4}

The experiments and LLM training were conducted on an AWS \emph{G5.xlarge} instance, equipped with 4 CPUs, 16 GB of RAM, and 1 GPU with 24 GB of memory. The consistent environment across all experiments ensures reproducibility and reliability of results. 

The AdamW optimizer facilitated the optimization of our LLM. AdamW extends the Adam optimizer, by directly incorporating weight decay into the update step. Weight decay is a regularization technique that penalizes large parameter values to prevent overfitting. The "W" in AdamW specifically stands for "Weight Decay," highlighting its integration of this regularization term. We set the learning rate to 2e-5., with a weight decay hyperparameter of 0.01. The beta coefficients are 0.9 and 0.999 for the gradient and its square respectively. 

\subsection{Failure Detection Algorithm (FDA)}\label{sub_4_4}

Our Masked Language Model (MLM) provided predictions at the chunk level, which we aggregated to the PCAP level for final failure detection. We computed two metrics per chunk: the Number of Misclassifications (NOM) and the Mean Negative Log-Likelihood (MNLL) loss. For each PCAP, we selected the top k chunks based on these metrics and calculated their average. This process aims to identify whether any chunks within a PCAP file exhibit signs of failure. In our experiments, we set the k hyperparameter to 3, using these metrics to determine the presence of failures in PCAP files. 

For the threshold-based FDA, we calculated the mean and standard deviation of NOM-k across the training set. PCAPs with a mean NOM-k exceeding three standard deviations from the mean were classified as containing failures, encompassing 99.7\% of PCAPs from the training set deemed successful. 

In the unsupervised ML-based approach to FDA, we employed the Elliptic Envelope (EE) algorithm. This method computes the covariance matrix and defines an ellipsoid around the central point of the data, presumed to follow a Gaussian distribution. It applies the Mahalanobis distance to identify outliers, which, in our context, are failed PCAPs. We leveraged both NOM-k and MNLL-k metrics as features for representing PCAP files. After training the MLM, we collected these metrics from the training data to inform the EE algorithm, which was then integrated as the concluding step in our pipeline. 

Upon receiving a new PCAP, it was first processed through the MLM to calculate the relevant metrics. Subsequently, the FDA was applied, using either the threshold-based or the EE algorithm. It is noteworthy that various other aggregation strategies, such as mean NOM or mean MNLL, were explored but did not yield satisfactory results. 

\section{Results and Discussion}\label{section_results}

In this section, we evaluate the impact of the PCAP file representation and the failure detection algorithms (\ref{sub_5_1}) on the performance of failure detection. We also assess the failure detection capability of LLMcap on two external datasets (\ref{sub_5_2}) and demonstrate an example of tagging relevant chunks when a failure is detected (\ref{sub_5_3}). Finally, we investigate the computational time for training and inference (\ref{sub_5_4}) of different approaches.  

\subsection{Best Representation and Best Failure Detection Algorithm}\label{sub_5_1}

\begin{figure}
\centerline{\includegraphics[width=\columnwidth]{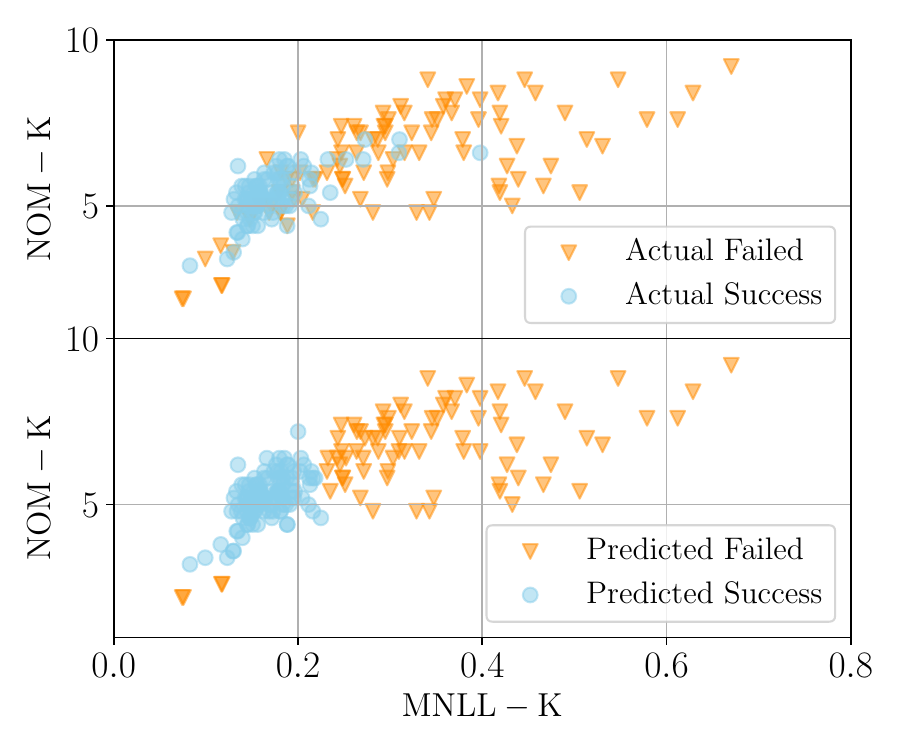}}
\caption{The visual representation of two measures, i.e., the number of misclassifications NOM-K and mean NLL-K loss, for ground-truth labels (top), and predictions of our method (bottom).}
\label{fig4}
\end{figure}

A series of experiments were conducted to determine the best PCAP file representation and the most effective Failure Detection algorithm. Our training dataset for the VoLTE service comprised 1311 successful PCAPs with subsample validation and test splits of 100 each. Additionally, a control set with 364 failed VoLTE PCAPs, not used during training was included. This resulted in 1111 successful PCAP files in the training set, 100 successful in the validation set, 100 successful in the test set, 100 with failure in the validation set, and 100 PCAP files with failure in the test set. In the validation and test sets, we kept the number of examples balanced between the two classes. Batch size and chunk size remained constant across the experiments with a batch size of 2 and each chunk containing 64 tokens. Figure \ref{fig4} illustrates the effectiveness of employed metrics, in filtering out failed PCAPs (orange triangles). Notably, successful PCAPs (blue circles), cluster together exhibiting lower misclassification and NLL loss values (Figure \ref{fig4}).

The results of our experiments are summarized in Table \ref{tab2}, presenting precision, recall, F1-score, and F2-score independently for each class. While reporting these metrics, the primary focus of LLMcap is to detect failure emphasizing recall over precision. For that reason, our objective criterion for evaluating is the F2-score on the failure class. In the first experiment examining the impact on PCAP representation, the optimal approach is to use key-value pairs (rows 1 and 2 of Table \ref{tab2}). Subsequently, we tested a change in the failure detection method, employing the EE algorithm, (rows 2 and 3 of Table \ref{tab2}). The results indicate that the EE algorithm is the superior failure detection method, improving the F2 score, although it does impact the precision of the failure class.  

The motivation behind employing the EE algorithm is to cover non-square differences between both classes for the utilized metrics. Furthermore, it allows us to combine NOM and MNLL loss, to fine-tune the model's accuracy (see Figure \ref{fig4}).

In the previous experiments, the DICT representation lacked separation across the packet level. To evaluate the impact of introducing such separation, we modified the method to create a vocabulary for each packet, later concatenating them into one. This approach differs from the previous one in the ordering of the tokens in the input sequence, thus changing the context for the LLM during MLM. We hypothesized that this alteration may improve the generalization capabilities by providing more consistent predictions and reducing the number of false positives and negatives. This approach, labeled as PCT-DICT with a threshold failure detection model and the EE model (PCT-DICT, rows 4 and 5 in Table \ref{tab2}), resulted in an increased F2 score. However, this representation in addition to the EE algorithm led to a lower precision for the Failure class. Consequently, we determined that the best combination is to use a reduced set of ordered key-value representations for the PCAP file alongside a threshold-based approach. In subsequent experiments, we exclusively retained the PCT-DICT-T model. In this scenario, the context window encapsulates packet-level information, enhancing reconstruction abilities and providing more consistent results due to the high correlation within the packet-level context. 

\begin{table*}[]
\centering
\begin{tabular}{|c|c|cccc|cccc|}
\hline
   \multirow{3}{*}{\centering Row \#}    & \multicolumn{1}{c|}{Class $\longrightarrow$}                               & \multicolumn{4}{c|}{Success Class}                                                                & \multicolumn{4}{c|}{Failure Class}                                                                \\ \cline{2-10}
 & \begin{tabular}[c]{@{}c@{}}Metric $\longrightarrow$ \\ Model $\downarrow$ \end{tabular} & \multicolumn{1}{c|}{Precision} & \multicolumn{1}{c|}{Recall} & \multicolumn{1}{c|}{F1}    & F2    & \multicolumn{1}{c|}{Precision} & \multicolumn{1}{c|}{Recall} & \multicolumn{1}{c|}{F1}    & F2    \\ \hline
1   & TEXT-T                                                   & \multicolumn{1}{c|}{0.688}     & \multicolumn{1}{c|}{\textbf{0.950}}  & \multicolumn{1}{c|}{0.798} & \textbf{0.882} & \multicolumn{1}{c|}{\textbf{0.919}}     & \multicolumn{1}{c|}{0.570}  & \multicolumn{1}{c|}{0.703} & 0.616 \\ 
2      & DICT-T                                                   & \multicolumn{1}{c|}{0.719}     & \multicolumn{1}{c|}{0.820}  & \multicolumn{1}{c|}{0.766} & 0.797 & \multicolumn{1}{c|}{0.790}     & \multicolumn{1}{c|}{0.680}  & \multicolumn{1}{c|}{0.731} & 0.699 \\ 
3      & DICT-EE                                                  & \multicolumn{1}{c|}{0.808}     & \multicolumn{1}{c|}{0.760}  & \multicolumn{1}{c|}{0.783} & 0.769 & \multicolumn{1}{c|}{0.773}     & \multicolumn{1}{c|}{0.820}  & \multicolumn{1}{c|}{0.796} & 0.810 \\
4      & PCT-DICT-T                                               & \multicolumn{1}{c|}{\textbf{0.857}}     & \multicolumn{1}{c|}{0.840}  & \multicolumn{1}{c|}{\textbf{0.848}} & 0.843 & \multicolumn{1}{c|}{0.843}     & \multicolumn{1}{c|}{0.860}  & \multicolumn{1}{c|}{\textbf{0.851}} & \textbf{0.857} \\
5      & PCT-DICT-EE                                              & \multicolumn{1}{c|}{0.850}     & \multicolumn{1}{c|}{0.570}  & \multicolumn{1}{c|}{0.682} & 0.610 & \multicolumn{1}{c|}{0.676}     & \multicolumn{1}{c|}{\textbf{0.900}}  & \multicolumn{1}{c|}{0.772} & 0.844 \\ \hline
\end{tabular}
\caption{Experimental results for different combinations of PCAP file representations and Failure Detection algorithm. TEXT and DICT are when the PCAP file representation is either full text or a set of selected key values respectively. T and EE denote the usage of the threshold-based approach and the Elliptic Envelope approach respectively. PCT denotes the usage of an ordered list of key values for PCAP file representation.}
\label{tab2}
\end{table*}

\subsection{Performances on External Services}\label{sub_5_2}

The subsequent experiments aimed to assess if a model trained on PCAP files from one service could effectively detect failures in other services. While the training phase utilized the Voice over LTE (VoLTE) service dataset, evaluations were conducted on other services, such as 4G-5G NSA Connectivity and Voice over New Radio (VoNR) datasets (see Table \ref{tab3}). The latter two services were not part of the training set and solely underwent the inference process. For the 4G-5G NSA Connectivity service, we collected 982 successful and 264 failed PCAPs. For the VoNR service, the dataset comprised 66 successful and 331 failed PCAPs. Results are presented in Table \ref{tab3}.  

The VoNR service dataset yielded comparable results to those obtained with the VoLTE dataset. The model achieved a higher precision for the failure class but at the cost of a lower recall. The nearly identical F1 scores for both classes suggest that the model performs well on this unseen dataset and adjustments to the threshold could enhance the F2 score. However, the model did not perform well on the 4G-5G-NSA Connectivity dataset, with the F2 score for the failure class dropping to 0.342. The difference in performances can be attributed to the similarity between VoLTE and VoNR services, sharing identical message flows and sequences, particularly in SIP, utilizing the same IMS components. Conversely, 4G-5G-NSA Connectivity scenarios have distinct message flows and sequences with a completely different structure, posing a challenge for the trained model that may not have learned this new syntax. While LLMcap demonstrated the ability to generalize across closely-related services without retraining, this generalization is limited when the services differ drastically. This limitation might be overcome through retraining or fine-tuning on external services. 

\begin{table*}[]
\centering
\begin{tabular}{|c|cccc|cccc|}
\hline
Class  $\longrightarrow$                 & \multicolumn{4}{c|}{Success Class}                                                                & \multicolumn{4}{c|}{Failure Class}                                                                \\ \cline{2-9}
Service    $\downarrow$            & \multicolumn{1}{c|}{Precision} & \multicolumn{1}{c|}{Recall} & \multicolumn{1}{c|}{F1}    & F2    & \multicolumn{1}{c|}{Precision} & \multicolumn{1}{c|}{Recall} & \multicolumn{1}{c|}{F1}    & F2    \\ \hline
VoNR                   & \multicolumn{1}{c|}{0.774}     & \multicolumn{1}{c|}{0.982}  & \multicolumn{1}{c|}{0.866} & 0.932 & \multicolumn{1}{c|}{0.980}     & \multicolumn{1}{c|}{0.753}  & \multicolumn{1}{c|}{0.852} & 0.790 \\ 
4G-5G NSA Connectivity & \multicolumn{1}{c|}{0.747}     & \multicolumn{1}{c|}{0.441}  & \multicolumn{1}{c|}{0.554} & 0.480 & \multicolumn{1}{c|}{0.177}     & \multicolumn{1}{c|}{0.447}  & \multicolumn{1}{c|}{0.254} & 0.342 \\ \hline
\end{tabular}
\caption{Experimental results for PCT-DICT-T model, when performing inference over other services, i.e., VoNR and 4G-5G NSA Connectivity.}
\label{tab3}
\end{table*}

\subsection{Tagging PCAP Content}\label{sub_5_3}

LLMcap offers the advantage of providing additional insights when a failure is detected in a PCAP file. By selecting chunks with high NOM, we can identify the reason for the classification. Figure \ref{fig5} illustrates an example of LLMcap output, where the model made 8 mispredictions in a chunk corresponding to frame 223. The highlighted bold section indicates where the errors occurred. In this scenario, users can examine the identified error, such as “500 Server Internal Error” to understand the cause of the failure. This feature enhances the interpretability of LLMcap’s findings, aiding users in diagnosing and addressing network issues. Reference for the error: RFC 3261, section 21.5.1.\cite{b13}. 

\begin{figure}[h]
    \centering
    \begin{adjustbox}{width=\columnwidth}
        \texttt{\begin{tabular}{|l|}
            \hline
            Found failure for PCAP: \\
            PCAP filename: VoLTE\_145.pcap, in frame: \textbf{223}. \\
            Number of misclassifications in chunk: 8. \\
            Chunk content: \\
            \ldots \\
            printable \_ value: 9 \\
            tcp. stream. printable \_ name: \\
            stream index: 9 \\
            geninfo. printable \_ name: general information \\
            \textbf{sip. status - line. printable \_ value: sip/2.0 500 }\\
            \textbf{server internal error - 20050 no available pool as for}\\ \textbf{triggered service found }\\
            dstport. printable \_ name: destination port: 37735 \\
            tcp. dstport. printable \_ value: 377 \\
            \ldots \\
            \hline
        \end{tabular}}
    \end{adjustbox}
    \caption{Listing shows an example of output from LLMcap. In this case, the model found 2 sections of SIP-related errors.}
    \label{fig5}
\end{figure}

\subsection{Execution Time}\label{sub_5_4}

In each experiment, we evaluated both the training and inference times (Table \ref{tab4}). The choice of PCAP file representation significantly influences the training and inference times. Using a full-text representation increases the number of chunks per file compared with a key-value representation. Notably, the ordering and the FDA have minimal impact on computational time.  

The parallelizability of LLMcap extends beyond a mere technical advantage; it underpins a scalable and efficient approach to network failure detection.  LLMcap is designed to promptly predict failures in individual data chunks, allowing for parallel processing across distributed computing resources. This asynchronous aggregation of results prevents network analysis from becoming a bottleneck, promoting timely failure detection even in telecommunication infrastructures.  

Furthermore, LLMcap has a minimal computational footprint, making it ideal for deployment in edge computing scenarios. Its efficiency and compact size are crucial for telecommunication networks where data processing closer to the source (at the edge) reduces latency, enhances data privacy, and lowers bandwidth usage by avoiding the need to transmit large volumes of PCAP data to centralized cloud-based systems.  LLMcap’s design not only meets the immediate need for accurate PCAP failure detection but also aligns with the broader objectives in modern network management, which increasingly favors decentralized and edge-focused computing solutions. 

\begin{table}[]
\centering
\begin{tabular}{|c|c|c|}
\hline
\begin{tabular}[c]{@{}c@{}}Time $\longrightarrow$ \\ Model $\downarrow$ \end{tabular}   & Training time {[}Hours{]}  & \begin{tabular}[c]{@{}c@{}} Average inference  \\ time per PCAP {[}Seconds{]}  \end{tabular} \\ \hline
\multicolumn{1}{|c|}{TEXT}    & $\sim$156.0          & 109.2       \\ \hline
\multicolumn{1}{|c|}{DICT}    & \multirow{4}{*}{$\sim$50.4} & 15.0  \\
\multicolumn{1}{|c|}{DICT-EE} &                            & 14.7 \\
PCT-DICT                      &                            & 10.0 \\
PCT-DICT-EE                   &                            & 9.5 \\ \hline
\end{tabular}
\caption{The training and inference time for each approach. The TEXT method utilizes all PCAP data, thus takes the longest time.}
\label{tab4}
\end{table}


\section{Conclusions}\label{section_conclusions}

We introduced LLMcap as an innovative approach to utilizing LLMs for PCAP file analysis. Operating without labeled data demonstrated high accuracy in failure detection and localization within PCAP data. Notably, LLMcap offers adaptability through interchangeable components without sacrificing performance. We proposed a PCAP data reduction into a key-value dictionary format pre-training, showcasing efficiency without significant quality loss in failure detection. Our findings highlighted the balance between context length and processing speeds in LLMs, with potential improvements in model understanding by extending context. Data filtering emerged as a beneficial strategy to enhance the model. LLMcap serves as a foundational framework, enabling failure detection and root cause analysis in production, while leveraging source PCAPs directly instead of traditional methods such as Performance and Configuration Management, or Key Performance Indicators. We also demonstrated that a key-value pair representation is optimal, indicating efficiency and simplicity in understanding PCAP language. We aim to extend our model's training set to incorporate a wider range of services to increase our approach's applicability. In conclusion, this research represents an advancement in PCAP analysis, unlocking the potential for real-time failure detection in live telecom networks. Our approach alleviates the challenges of traditional troubleshooting, paving the way for future developments in network quality of service, fault detection, and performance optimization using advanced language models.


\vspace{12pt}
\end{document}